\documentclass{article}

\usepackage{microtype}
\usepackage{graphicx}
\usepackage{subfigure}
\usepackage{booktabs} 

\usepackage{hyperref}
\usepackage{graphicx} 
\usepackage{amsmath} 
\usepackage{amssymb}  
\usepackage{cite}
\usepackage{flexisym}

\usepackage[accepted]{icml2020}

\icmltitlerunning{Learning Object-Centric Navigation Policies on Semantic Maps with Graph Convolutional Networks}

\begin{document}

\twocolumn[
\icmltitle{Where are the Keys? -- Learning Object-Centric Navigation Policies on Semantic Maps with Graph Convolutional Networks}




\begin{icmlauthorlist}
\icmlauthor{Niko Suenderhauf}{QUT}
\end{icmlauthorlist}

\icmlaffiliation{QUT}{Queensland University of Technology (QUT) Centre for Robotics, Brisbane, Australia. This work was supported
by the Australian Research Council Centre of Excellence for Robotic Vision
under project CE140100016.\\ \texttt{niko.suenderhauf@qut.edu.au}}

\icmlkeywords{}
\vskip 0.3in
]



\printAffiliationsAndNotice{}  

\newcommand{\vect}[1]{\mathbf{ #1}}
\newcommand{\vectg}[1]{{\boldsymbol{ #1}}}
\newcommand{\ggo}{\ensuremath{\mathrm{g^2o}} }
\newcommand{\R}{\mathbb{R}}
\newcommand{\N}{\mathbb{N}}
\newcommand{\Z}{\mathbb{Z}}
\renewcommand{\P}{\mathbb{P}}
\newcommand{\tran}{^\top}
\newcommand{\T}{^\mathsf{T}}
\newcommand{\iT}{^{-\mathsf{T}}}
\newcommand{\inv}{^{-1}}
\newcommand{\func}[2]{\mathtt{#1}\left\{#2\right\}}
\newcommand{\sig}{\operatorname{sig}}
\newcommand{\diag}{\operatorname{diag}}
\newcommand{\argmin}{\operatornamewithlimits{argmin}}
\newcommand{\argmax}{\operatornamewithlimits{argmax}}
\newcommand{\RMSE}{\operatorname{RMSE}}
\newcommand{\RMSEpos}{\operatorname{RMSE}_\text{pos}}
\newcommand{\RMSEori}{\operatorname{RMSE}_\text{ori}}
\newcommand{\RPE}{\operatorname{RPE}}
\newcommand{\RPEpos}{\operatorname{RPE}_\text{pos}}
\newcommand{\RPEori}{\operatorname{RPE}_\text{ori}}
\newcommand{\rpe}{\varepsilon_{\vdelta}}
\newcommand{\achiError}{\bar{e}_{\chi^2}}
\newcommand{\chiError}{e_{\chi^2}}
\newcommand{\normal}[2]{\mathcal{N}\left(#1, #2\right)}
\newcommand{\uniform}[2]{\mathcal{U}\left(#1, #2\right)}
\newcommand{\pfrac}[2]{\frac{\partial #1}{\partial #2}}  
\newcommand{\fracpd}[2]{\frac{\partial #1}{\partial #2}} 
\newcommand{\fracppd}[2]{\frac{\partial^2 #1}{\partial #2^2}}  
\newcommand{\dd}{\mathrm{d}}  
\newcommand{\smd}[2]{\left\| #1 \right\|^2_{#2}}
\newcommand{\E}[1]{\text{\normalfont{E}}\left[ #1 \right]}     
\newcommand{\Cov}[1]{\text{\normalfont{Cov}}\left[ #1 \right]} 
\newcommand{\Var}[1]{\text{\normalfont{Var}}\left[ #1 \right]} 
\newcommand{\Tr}[1]{\text{\normalfont{tr}}\left( #1 \right)}   
\def\sgn{\mathop{\mathrm sgn}}    
\newcommand{\twovector}[2]{\begin{pmatrix} #1 \\ #2 \end{pmatrix}} 
\newcommand{\smalltwovector}[2]{\left(\begin{smallmatrix} #1 \\ #2 \end{smallmatrix}\right)} 
\newcommand{\threevector}[3]{\begin{pmatrix} #1 \\ #2 \\ #3 \end{pmatrix}} 
\newcommand{\fourvector}[4]{\begin{pmatrix} #1 \\ #2 \\ #3 \\ #4 \end{pmatrix}}  
\newcommand{\smallthreevector}[3]{\left(\begin{smallmatrix} #1 \\ #2 \\ #3 \end{smallmatrix}\right)} 
\newcommand{\fourmatrix}[4]{\begin{pmatrix} #1 & #2 \\ #3 & #4 \end{pmatrix}} 
\newcommand{\vA}{\vect{A}}
\newcommand{\vB}{\vect{B}}
\newcommand{\vC}{\vect{C}}
\newcommand{\vD}{\vect{D}}
\newcommand{\vE}{\vect{E}}
\newcommand{\vF}{\vect{F}}
\newcommand{\vG}{\vect{G}}
\newcommand{\vH}{\vect{H}}
\newcommand{\vI}{\vect{I}}
\newcommand{\vJ}{\vect{J}}
\newcommand{\vK}{\vect{K}}
\newcommand{\vL}{\vect{L}}
\newcommand{\vM}{\vect{M}}
\newcommand{\vN}{\vect{N}}
\newcommand{\vO}{\vect{O}}
\newcommand{\vP}{\vect{P}}
\newcommand{\vQ}{\vect{Q}}
\newcommand{\vR}{\vect{R}}
\newcommand{\vS}{\vect{S}}
\newcommand{\vT}{\vect{T}}
\newcommand{\vU}{\vect{U}}
\newcommand{\vV}{\vect{V}}
\newcommand{\vW}{\vect{W}}
\newcommand{\vX}{\vect{X}}
\newcommand{\vY}{\vect{Y}}
\newcommand{\vZ}{\vect{Z}}
\newcommand{\va}{\vect{a}}
\newcommand{\vb}{\vect{b}}
\newcommand{\vc}{\vect{c}}
\newcommand{\vd}{\vect{d}}
\newcommand{\ve}{\vect{e}}
\newcommand{\vf}{\vect{f}}
\newcommand{\vg}{\vect{g}}
\newcommand{\vh}{\vect{h}}
\newcommand{\vi}{\vect{i}}
\newcommand{\vj}{\vect{j}}
\newcommand{\vk}{\vect{k}}
\newcommand{\vl}{\vect{l}}
\newcommand{\vm}{\vect{m}}
\newcommand{\vn}{\vect{n}}
\newcommand{\vo}{\vect{o}}
\newcommand{\vp}{\vect{p}}
\newcommand{\vq}{\vect{q}}
\newcommand{\vr}{\vect{r}}
\newcommand{\vs}{\vect{s}}
\newcommand{\vt}{\vect{t}}
\newcommand{\vu}{\vect{u}}
\newcommand{\vv}{\vect{v}}
\newcommand{\vw}{\vect{w}}
\newcommand{\vx}{\vect{x}}
\newcommand{\vy}{\vect{y}}
\newcommand{\vz}{\vect{z}}
\newcommand{\valpha}{\vectg{\alpha}}
\newcommand{\vbeta}{\vectg{\beta}}
\newcommand{\vgamma}{\vectg{\gamma}}
\newcommand{\vdelta}{\vectg{\delta}}
\newcommand{\vepsilon}{\vectg{\epsilon}}
\newcommand{\vtau}{\vectg{\tau}}
\newcommand{\vmu}{\vectg{\mu}}
\newcommand{\vphi}{\vectg{\phi}}
\newcommand{\vPhi}{\vectg{\Phi}}
\newcommand{\vpi}{\vectg{\pi}}
\newcommand{\vPi}{\vectg{\Pi}}
\newcommand{\vPsi}{\vectg{\Psi}}
\newcommand{\vchi}{\vectg{\chi}}
\newcommand{\vvarphi}{\vectg{\varphi}}
\newcommand{\veta}{\vectg{\eta}}
\newcommand{\viota}{\vectg{\iota}}
\newcommand{\vkappa}{\vectg{\kappa}}
\newcommand{\vlambda}{\vectg{\lambda}}
\newcommand{\vLambda}{\vectg{\Lambda}}
\newcommand{\vnu}{\vectg{\nu}}
\newcommand{\vgo}{\vectg{\o}}
\newcommand{\vvarpi}{\vectg{\varpi}}
\newcommand{\vtheta}{\vectg{\theta}}
\newcommand{\vvartheta}{\vectg{\vartheta}}
\newcommand{\vTheta}{\vectg{\Theta}}
\newcommand{\vrho}{\vectg{\rho}}
\newcommand{\vsigma}{\vectg{\sigma}}
\newcommand{\vSigma}{\vectg{\Sigma}}
\newcommand{\vvarsigma}{\vectg{\varsigma}}
\newcommand{\vupsilon}{\vectg{\upsilon}}
\newcommand{\vomega}{\vectg{\omega}}
\newcommand{\vOmega}{\vectg{\Omega}}
\newcommand{\vxi}{\vectg{\xi}}
\newcommand{\vXi}{\vectg{\Xi}}
\newcommand{\vpsi}{\vectg{\psi}}
\newcommand{\vzeta}{\vectg{\zeta}}
\newcommand{\vzero}{\vect{0}}
\newcommand{\cA}{\mathcal{A}}
\newcommand{\cB}{\mathcal{B}}
\newcommand{\cC}{\mathcal{C}}
\newcommand{\cD}{\mathcal{D}}
\newcommand{\cE}{\mathcal{E}}
\newcommand{\cF}{\mathcal{F}}
\newcommand{\cG}{\mathcal{G}}
\newcommand{\cH}{\mathcal{H}}
\newcommand{\cI}{\mathcal{I}}
\newcommand{\cJ}{\mathcal{J}}
\newcommand{\cK}{\mathcal{K}}
\newcommand{\cL}{\mathcal{L}}
\newcommand{\cM}{\mathcal{M}}
\newcommand{\cN}{\mathcal{N}}
\newcommand{\cO}{\mathcal{O}}
\newcommand{\cP}{\mathcal{P}}
\newcommand{\cQ}{\mathcal{Q}}
\newcommand{\cR}{\mathcal{R}}
\newcommand{\cS}{\mathcal{S}}
\newcommand{\cT}{\mathcal{T}}
\newcommand{\cU}{\mathcal{U}}
\newcommand{\cV}{\mathcal{V}}
\newcommand{\cW}{\mathcal{W}}
\newcommand{\cX}{\mathcal{X}}
\newcommand{\cY}{\mathcal{Y}}
\newcommand{\cZ}{\mathcal{Z}}
\newcommand{\fA}{\mathfrak{A}}
\newcommand{\fB}{\mathfrak{B}}
\newcommand{\fC}{\mathfrak{C}}
\newcommand{\fD}{\mathfrak{D}}
\newcommand{\fE}{\mathfrak{E}}
\newcommand{\fF}{\mathfrak{F}}
\newcommand{\fG}{\mathfrak{G}}
\newcommand{\fH}{\mathfrak{H}}
\newcommand{\fI}{\mathfrak{I}}
\newcommand{\fJ}{\mathfrak{J}}
\newcommand{\fK}{\mathfrak{K}}
\newcommand{\fL}{\mathfrak{L}}
\newcommand{\fM}{\mathfrak{M}}
\newcommand{\fN}{\mathfrak{N}}
\newcommand{\fO}{\mathfrak{O}}
\newcommand{\fP}{\mathfrak{P}}
\newcommand{\fQ}{\mathfrak{Q}}
\newcommand{\fR}{\mathfrak{R}}
\newcommand{\fS}{\mathfrak{S}}
\newcommand{\fT}{\mathfrak{T}}
\newcommand{\fU}{\mathfrak{U}}
\newcommand{\fV}{\mathfrak{V}}
\newcommand{\fW}{\mathfrak{W}}
\newcommand{\fX}{\mathfrak{X}}
\newcommand{\fY}{\mathfrak{Y}}
\newcommand{\fZ}{\mathfrak{Z}}

\begin{abstract}
Emerging object-based SLAM algorithms can build a graph representation of an environment comprising nodes for robot poses and object landmarks. However, while this map will contain static objects such as furniture or appliances, many moveable objects (e.g. the \emph{car keys}, the \emph{glasses}, or a \emph{magazine}), are not suitable as landmarks and will not be part of the map due to their non-static nature. We show that Graph Convolutional Networks can learn navigation policies to find such unmapped objects by learning to exploit the hidden probabilistic model that governs where these objects appear in the environment. The learned policies can generalise to object classes unseen during training by using word vectors that express semantic similarity as representations for object nodes in the graph. Furthermore, we show that the policies generalise to unseen environments with only minimal loss of performance. We demonstrate that pre-training the policy network with a proxy task can significantly speed up learning, improving sample efficiency. 
\end{abstract}

\section{Introduction}
Where should a domestic service robot go look for the misplaced keys? Where would it find a bottle of milk, where would the owners have placed the remote control for the TV? Humans have an intuitive understanding of where to successfully search for these objects: The remote will most likely be found in the vicinity of the TV, the sofa, or the armchair. The milk will most likely be in the fridge, but maybe someone left it out on the kitchen table or the benchtop. In human-made environments, many -- if not most -- objects are not placed randomly, but tend to appear in proximity to a small set of other objects with related semantics or functionality. Humans have intuitive access to this underlying probabilistic process. 

\begin{figure}[t]
    \centering
    \includegraphics[width=1\linewidth]{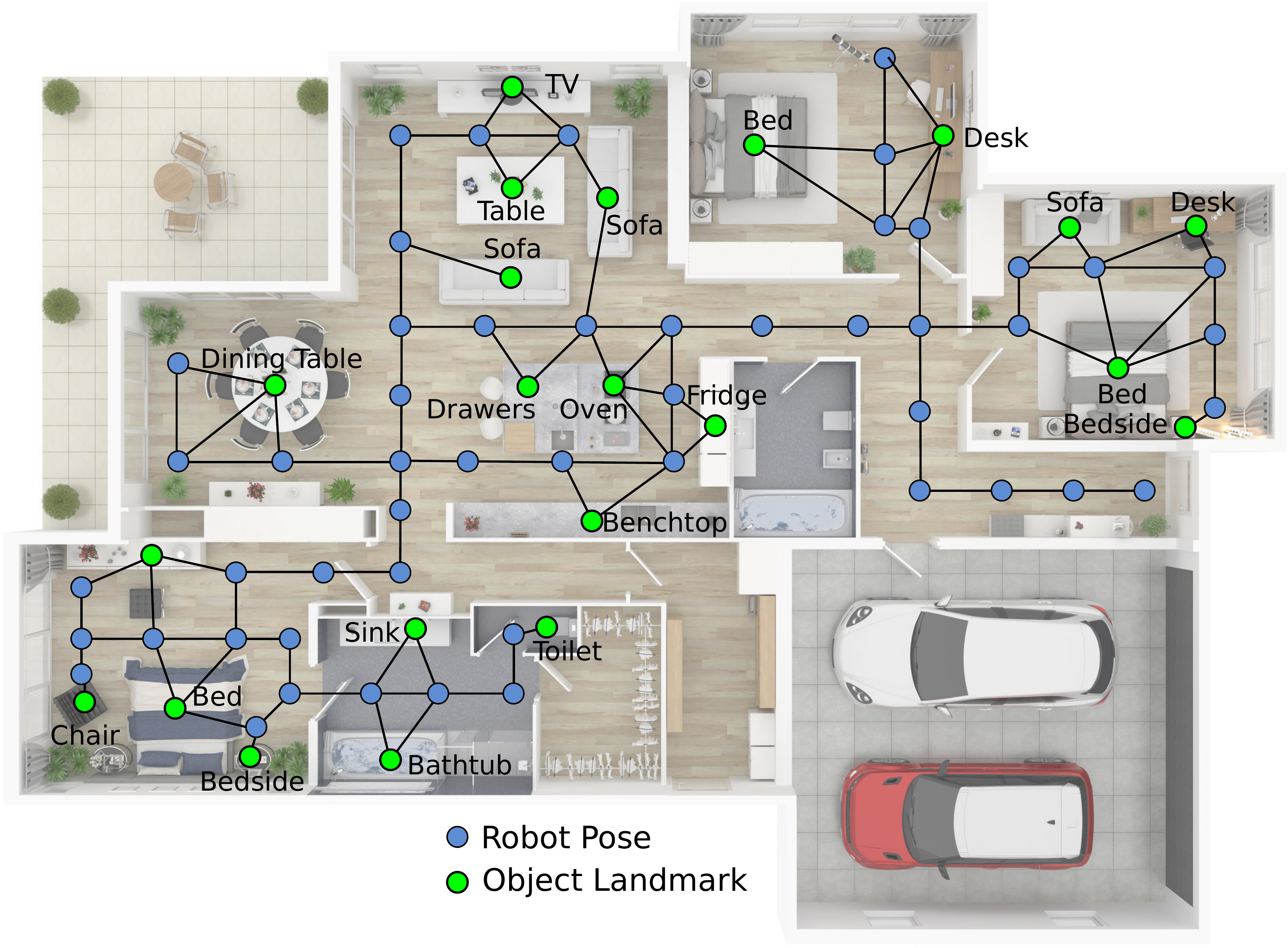}
    \caption{Given a graph-based semantic map with pose nodes and \emph{static} object landmark nodes, where should a domestic service robot look for a \emph{non-static} target object that would not appear in the map, such as the \emph{keys} or the TV \emph{remote}? We propose to learn a navigation policy for this task by training a graph convolutional network with reinforcement learning. \vspace{-1em}
    }
    \label{fig:main_idea}
\end{figure}

In this paper, we investigate if an agent can learn a navigation policy on a graph-based map comprised of pose nodes and \emph{static} object landmarks (see Fig.~\ref{fig:main_idea} for an illustration). We show that a graph convolutional network can act as a policy network, providing a distribution over the pose nodes in the graph map. By training this network with reinforcement learning and sampling navigation goals from the resulting distribution, an agent can learn to find objects that do \emph{not} appear in the map due to their \emph{non-static} nature. 

In contrast to recent work that learns to navigate directly from raw images, our approach learns on the graph maps constructed by an emerging family of object-based semantic SLAM systems~\cite{nicholson2018quadric,hosseinzadeh2019real,hosseinzadeh2018structure,hu2019deepslam,zhi2019scenecode,salasMoreno2013slam++,McCormac2018}. While approaches learning from raw pixels typically take in the order of millions of episodes to train, our approach learns much faster, converging after a few 10,000 episodes. We present a method for pre-training based on a proxy task that speeds up the learning process even further. 

As we will demonstrate, the learned policies can generalise to objects never encountered during training. This is possible because we use FastText word vectors~\cite{bojanowski2016fasttext} to represent object landmarks in the graph map. These vector representations have been trained on a large corpus of text data and capture semantic similarity between different words. If during deployment the target object is unknown (e.g. \emph{butter} or \emph{yoghurt}), but semantically similar to one of the known objects (e.g. \emph{milk}), and it tends to appear in similar places (e.g. in the \emph{fridge} or the \emph{kitchen table}), the policy has a high chance of successfully navigating towards it within the given time budget. We also show that the policies are independent of the graph structure and layout, i.e. they generalise to unknown environments with only a minimal drop in performance.


\section{Related Work}
\subsection{Object-based Semantic Mapping and SLAM}
Our research is motivated by the emerging body of work in \emph{semantic} Simultaneous Localisation and Mapping (SLAM)~\cite{cadena2016past} that focuses on building semantically rich representations of an environment, with \emph{objects} as the central entities in the resulting map. After early approaches such as SLAM++~\cite{salasMoreno2013slam++} introduced the idea of object-based SLAM, recent advances in deep learning for visual object detection and semantic segmentation enabled a range of object-based SLAM systems to appear. QuadricSLAM~\cite{nicholson2018quadric} and others~\cite{hosseinzadeh2019real,hosseinzadeh2018structure} investigated the utility of representing objects with low-dimensional geometric primitives, whereas Fusion++~\cite{McCormac2018} represents objects as Truncated Signed Distance Functions. The expressive power of latent code representations for SLAM has been investigated in~\cite{zhi2019scenecode}. All these emerging object-centric SLAM techniques can build a rich graph-based representation of an environment, containing pose nodes and object nodes. Our approach for learning navigation policy processes this graph-based map with a graph convolutional network.


\subsection{Object Goal-directed Navigation}
Our work is concerned with learning a policy that enables an agent or robot to find a target object of a specific class in its environment, corresponding to the ObjectGoal task in the taxonomy recently defined by~\cite{anderson2018evaluation}. Notice that since the target location is not known, classical path planning approaches are not applicable.
Recently there has been a lot of interest in \emph{visual} navigation, i.e. navigating to a goal without using an explicit map, but basing all decisions on visual input data. Different approaches attempt to solve this problem by using raw image data~\cite{zhu2017target, mirowski2018learning,mirowski2016learning,bruce2018learning}, mid-level features~\cite{sax2018mid} or high-level features such as segmentation masks~\cite{mousavian2019visual}, while others build implicit map representations to aid navigation~\cite{Gupta_2017_CVPR,savinov2018semi}.

Instead of attempting to learn goal-directed navigation policies end-to-end directly from visual data, we propose to leverage the growing body of work in object-based semantic SLAM to provide a representation of the environment. We show that effective navigation policies can be learned efficiently on this \emph{explicit} representation of the world with reinforcement learning and graph convolutional networks.





\subsection{Graph Neural Networks}
While the application of deep neural networks for image data has become ubiquitous in recent years, the more general concept of deep learning on graph data is not as widespread. 
With graph neural networks, nodes in a graph aggregate the feature vectors of neighbouring nodes to compute their own new feature vector. By applying this aggregation scheme $k$ times, a node's representation is a function of the information contained within the node's $k$-hop neighbourhood~\cite{xu2018powerful}. A comprehensive survey of graph neural networks is provided by~\cite{wu2019comprehensive}. In our work, we will use a Graph Convolutional Network~\cite{kipf2016semi} to aggregate semantic landmark representations within a graph-based map. We will train this network via reinforcement learning. While others have explored the application of reinforcement learning with graph neural networks for various tasks~\cite{mittal2019learning,madjiheurem2019representation,jiang2018graph,you2018graph,khan2019graph}, ours is the first work that investigates the efficacy of graph convolutional networks acting as policy networks for navigation on semantic graph-based maps. A very recent paper~\cite{nguyen2019reinforcement} used a semantic knowledge graph to aid navigation and proposed to use a graph convolutional network to extract an embedding of the semantic knowledge that was then concatenated with the representation of a target object and the visual input, and fed into a MLP-based policy network. In contrast, we propose to use the graph convolutional network as the policy network, training the policy directly on the high-level graphical representation of the environment, instead of a combination of semantic knowledge plus visual input.


\section{Problem Description and Assumptions}
\label{sec:assumptions}
We propose an approach to learn a navigation policy that enables a robot to find small, non-static household objects such as keys, or glasses, within a domestic environment. We make the following assumptions that are reasonable for a domestic service robot application: 

\noindent
(1) The robot has mapped the environment using an object-based semantic SLAM system such as~\cite{nicholson2018quadric} or others~\cite{hosseinzadeh2019real,hosseinzadeh2018structure,hu2019deepslam,zhi2019scenecode,salasMoreno2013slam++,McCormac2018}.
    
\noindent    
(2) The available semantic map is a graph structure, with pose nodes and object landmark nodes. An edge between two pose nodes represents that the robot can navigate from one pose to the other. An edge between an object landmark and a pose indicates the object is in range for useful interaction.

\noindent    
(3) Objects in the map are static\footnote{Current object-based semantic SLAM systems such as~\cite{nicholson2018quadric,hosseinzadeh2019real,hosseinzadeh2018structure,hu2019deepslam,zhi2019scenecode,salasMoreno2013slam++,McCormac2018} assume a static environment and cannot use dynamic objects as landmarks.} and will therefore be furniture items or appliances such as \emph{table}, \emph{bed}, or \emph{fridge}.
  
\noindent    
(4) The objects of interest that need to be found by the robot are smaller, moveable objects that are not mapped due to their non-static nature, such as \emph{keys}, \emph{glasses}, \emph{cup}, or \emph{remote}. 

\noindent    
(5) These objects of interest appear in the vicinity of the mapped objects with a certain probability. E.g. a \emph{remote control} will appear with some probability at the \emph{sofa}, \emph{armchair}, or \emph{TV}, but never in the \emph{wardrobe} or \emph{fridge}.

\noindent    
(6) The probabilistic model underlying this process is unknown and not directly accessible to the robot. Every time we evaluate the policy (and for every episode during training), the objects of interest are randomly placed in the environment (governed by the hidden probabilistic process).
    
\noindent    
(7) The task of the policy is to find an object of interest. In the following, we will refer to these objects of interest as \emph{target objects}. One policy should be capable of navigating to \emph{all} target objects, i.e. we do not learn a target-specific policy. 
 
\noindent
(8) The policy acts as a high-level planner, proposing to navigate to a pose node in the graph-based map. We assume that the robot has sufficient low-level navigation capabilities to reach this goal pose by a combination of path planning on the map and low-level motion control paired with reactive obstacle avoidance. We also assume the robot can localise itself with respect to the given map, building on the capabilities of current SLAM systems~\cite{nicholson2018quadric,hosseinzadeh2019real,hosseinzadeh2018structure,hu2019deepslam,zhi2019scenecode,salasMoreno2013slam++,McCormac2018}.

\begin{figure*}[t]
    \centering
    \includegraphics[width=\linewidth]{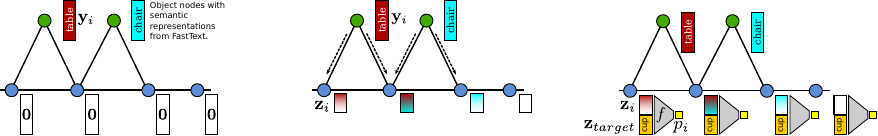}
    \caption{Concept of our policy network based on a Graph Convolutional Layer and 3 fully connected layers $f$. Left: We initialise the graph and represent each landmark with a word vector $\vy_i$ obtained from FastText~\cite{bojanowski2016fasttext}. Pose node representations are initialised as $\mathbf{0}$. Middle: A single graph convolutional layer propagates information from the landmark nodes into adjacent pose nodes, obtaining updated and compacted representations $\vz_i$. Right: The fully connected network $f$ obtains a distribution $p_i$ over all pose nodes from the representation of a non-mapped target object $\vz_{\text{target}}$ concatenated with the $\vz_i$.}
    \label{fig:concept}
\end{figure*}

\section{Approach}
We propose to learn a policy $\pi(\cG,c_{\text{target}})$ to find a target object of class $c_{\text{target}}$ in an environment represented by a graph-based map $\cG$. 
As we explained in Section \ref{sec:assumptions}, the target object is \emph{not} part of the map represented by $\cG$. Instead, the policy $\pi$ needs to learn the hidden probabilistic model that governs the complex relationship between the mapped objects and potential target objects.

We implement $\pi(\cG,c_{\text{target}})$ as a neural network, consisting of a Graph Convolutional Layer~\cite{kipf2016semi} and fully connected layers. This network acts as a policy network, providing a distribution over the pose nodes in $\cG$, conditioned on $c_{\text{target}}$. A navigation goal for the agent is selected by sampling from $\pi(\cG,c_{\text{target}})$. We train $\pi$ with REINFORCE, a simple policy gradient method~\cite{sutton2018reinforcement}. 

\subsection{Details of the Graph-based Map Representation $\cG$}
The policy network $\pi(\cG,c_{\text{target}})$ operates on a graph-based map $\cG = (\cX \cup \cL, \cE)$ that comprises robot pose nodes $\cX$, landmark nodes $\cL$, and edges $\cE$. 
Such a graph can be constructed easily from the outputs of a modern object-based SLAM system~\cite{nicholson2018quadric,hosseinzadeh2019real}. We assume the robot poses $\vx_i \in \cX$ are elements of SE3\footnote{Without loss of generality for our proposed method they could also be elements of SE2 or even $\R^2$ (representing only position).}. The landmarks $\vl_j \in \cL$ comprise geometric information such as their pose and shape, and a class label $c_j \in \cC^{\text{map}}$. The set $\cC^{\text{map}}$ comprises all object classes that can be mapped by the semantic SLAM algorithm that constructs $\cG$. As explained in Section \ref{sec:assumptions}, this will be \emph{static} objects such as furniture items and appliances. 

The target object class $c_{\text{target}}$ will be from the set $\cC^{\text{targets}}$ that contains non-static classes such as \emph{keys} or \emph{glasses} which will never appear as landmarks in the map $\cG$. We note that strictly $\cC^{\text{map}} \cap \cC^{\text{targets}} = \emptyset$ and also $c_{\text{target}} \notin \cC^{\text{map}}$.

\subsection{Augmenting the Graph Map with Word Vectors}
We augment the graph by adding a semantic representation in the form of a vector $\vy_i$ to every node. It is these representations $\vy_i$ that the Graph Convolutional Layer in $\pi$ will operate on. For a landmark node $\vl_i$ with class label $c_i$, we use the FastText~\cite{bojanowski2016fasttext} word vector corresponding to $c_i$ as its
semantic representation $\vy_i$. The FastText word vectors are continuous 300-dimensional representations and have been trained on a large corpus of text data comprising 16 billion tokens from Wikipedia, the UMBC webbase corpus and the statmt.org news dataset\footnote{Specifically, we used the pre-trained representations available in the \texttt{wiki-news-300d-1M.vec.zip} file from the FastText website \texttt{https://fasttext.cc/docs/en/english-vectors.html}.}. These word vectors maintain semantic similarity: two words with similar meaning will have vector representations that have a small cosine distance~\cite{bojanowski2016fasttext}.

Since pose nodes $\vx_i$ do not carry immediate semantic information, we initialise their respective semantic representation $\vy_i$ with $\mathbf{0}=(0,0, \dots, 0)^{\mathsf{T}}$.


\subsection{A Graph Convolutional Network as Policy Network}
We implement\footnote{We use PyTorch Geomtric~\cite{FeyLenssen2019} as the basis for our implementation.} the policy $\pi$ as a neural network consisting of one graph convolutional layer followed by three fully connected layers, with ReLu nonlinearities between all layers.

The graph convolutional layer implements the graph convolution operator proposed in~\cite{kipf2016semi}: 
\begin{equation}
    \vZ = \sigma(\hat\vD^{-\frac{1}{2}} \hat\vA \hat\vD^{-\frac{1}{2}}\vY\vTheta)     
    \label{eq:gconv}
\end{equation}
where $\vY \in \R^{N \times 300}$ is the matrix of all node representations $\vy_i$ of dimensionality 300, $\vTheta \in \R^{300\times 64}$ is the weight matrix of the graph convolutional layer, and $\sigma$ is the element-wise ReLu function. $\hat\vA = \vA + \vI$ denotes the graph adjacency matrix $\vA$ with inserted self-loops, and $\hat D_{ii} = \sum_j \vA_{ij}$ is the diagonal degree matrix of the graph. For each node in the graph, operation (\ref{eq:gconv}) essentially accumulates the representations of all the node's direct neighbours. It compresses the original 300-dimensional representations into a more compact 64-dimensional representation $\vz_i$. Fig.~\ref{fig:concept} illustrates the concept.

A network $f(\vz_i,, \vz_{\text{target}})$ of three fully connected layers $f_1, f_2, f_3$ calculates the final output $p_i$ of the policy network for each node in the graph. The 64-dimensional representation of the target class is obtained by multiplying the word vector representation of the target $\vy_{\text{target}}$ with the parameters of the graph convolutional layer: $\vz_{\text{target}} =  \sigma(\vy_{\text{target}} \cdot \vTheta$). Before passing it through the first layer $f_1$,  $\vz_{\text{target}}$ gets concatenated with the representation of each node $\vz_i$. The input and output dimensionalities of the three layers is (128, 64), (64, 32), (32, 1) respectively\footnote{Empiricially chosen. A comparative study on different network architectures or hyperparameters is beyond the scope of this paper.}. 

The final output of the policy network $\pi(\cG,c_{\text{target}})$ is a vector $\vp$ with $p_i$ acting as the logits to a categorical probability distribution over all nodes in the graph. By sampling from that distribution, we obtain a navigation goal for the agent\footnote{The described implementation does not distinguish between pose nodes and landmark nodes in the graph. Therefore, before sampling from the distribution $\pi$, we set the values of $p_i$ for all nodes corresponding to landmarks to $-100$, so that they are essentially never chosen as a goal.}. 

\subsection{Training the Graph Convolutional Policy Network}
We train the policy network  $\pi(\cG,c_{\text{target}})$ with REINFORCE, a simple policy gradient method. We use Adam as optimiser and set the initial learning rate to $10^{-4}$. In order to obtain a reward, the agent has to find the target object within 10 time steps, i.e. it can only select 10 navigation goals before the episode ends. We assume the agent can always reach the selected goal and do not take the geodesic distance to the selected goal into account. Future work could also reward the shortest path length.

To obtain a robust evaluation, we train 10 agents for 50,000 episodes. For increased robustness and to show that the specifics of the probabilistic model $\cP$ that governs the placement of target objects have no influence on the results, we repeat the training and evaluation in 20 environments with different underlying probabilistic models. Thus, in total we train 200 agents. 


\subsection{Datasets: Constructing Graph Maps for Training and Evaluation}
The graph maps used for training and evaluation are all constructed following the same principle: We first build the pose graph backbone, consisting of 1000 nodes and their connecting edges. Pose nodes are randomly assigned one of four \emph{room types} from the set (\emph{kitchen, bedroom, living room, office}). When a new pose node is inserted, it has a 95\% chance of being of the same room type  as its predecessor. 
After the pose graph backbone is built, we start populating the map with object landmarks. Every map contains between 100 and 500 objects from the set $\cC^{\text{map}}$. The exact number of objects in the map is randomly chosen from a uniform distribution. For every object that is to be placed, we randomly choose a pose node to connect it to (the object will be "visible" from that pose). A pose node can have multiple connected objects. Depending on the room type of the pose node, the object class is drawn from one of 20 classes\footnote{\emph{bed, bedside, wardrobe, cabinet, chair} appear in the \emph{bedroom}; \emph{kitchen-table, fridge, microwave, drawers, oven, cabinet, chair, benchtop} appear in the \emph{kitchen}; the \emph{living room} spawns objects of classes \emph{sofa, armchair, TV, dining-table}; and \emph{desk, shelf, chair} appear in the \emph{office}.}. To simulate objects being visible from more than one pose, we keep connecting the landmark to the next pose node with 50\% probability for each next pose, as long as the poses are of the same room type.

Target objects do not appear directly in the map. Instead, they spawn in the proximity of mapped objects according to a probabilistic process $\cP(c_{\text{target}} | c_{\text{map}})$. This probabilistic process $\cP$ is itself randomised in training and evaluation: the probability values are drawn from a uniform distribution between $0.1$ and $0.9$. However, target objects can only appear in the proximity of certain map objects\footnote{
\emph{(kitchen-table, benchtop, drawers, dining-table}) $ \rightarrow$ \emph{knife}. 
(\emph{kitchen-table, benchtop, drawers, dining-table}) $ \rightarrow$ \emph{fork}. 
(\emph{kitchen-table, benchtop, drawers, dining-table}) $ \rightarrow$ \emph{spoon}. 
(\emph{kitchen-table, benchtop, drawers, dining-table}) $ \rightarrow$ \emph{bowl}. 
(\emph{kitchen-table, benchtop, drawers, desk, dining-table}) $ \rightarrow$ \emph{cup}. 
(\emph{kitchen-table, benchtop, drawers, dining-table}) $ \rightarrow$ \emph{glass}. 
(\emph{kitchen-table, fridge}) $ \rightarrow$ \emph{milk}. 
(\emph{kitchen-table, fridge, dining-table}) $ \rightarrow$ \emph{beer}. 
(\emph{fridge}) $ \rightarrow$ \emph{apple}. 
(\emph{fridge}) $ \rightarrow$ \emph{juice}. 
(\emph{fridge}) $ \rightarrow$ \emph{oranges}. 
(\emph{bed, sofa}) $ \rightarrow$ \emph{pillow}. 
(\emph{bed, wardrobe, cabinet}) $ \rightarrow$ \emph{t-shirt}. 
(\emph{bed, wardrobe, cabinet}) $ \rightarrow$ \emph{pants}. 
(\emph{wardrobe, chair}) $ \rightarrow$ \emph{jacket}. 
(\emph{wardrobe, cabinet}) $ \rightarrow$ \emph{socks}. 
(\emph{bedside, desk, sofa, armchair}) $ \rightarrow$ \emph{glasses}. 
(\emph{bedside, desk, sofa, armchair, TV}) $ \rightarrow$ \emph{keys}. 
(\emph{bedside, shelf, sofa}) $ \rightarrow$ \emph{book}. 
(\emph{sofa, armchair, TV}) $ \rightarrow$ \emph{remote}. 
\label{note:spawn}}. For example, for a specific instantiation of $\cP$, a map object of class \emph{dining-table} will have a chance of $\cP(\text{knife} | \text{dining-table}) = 0.2$ of spawning a \emph{knife}. A different instantiation of $\cP$ might assign a probability of $0.5$.

\subsection{A Proxy Task for Pre-Training the Policy Network}
\label{sec:pretrain}
The graph convolutional layer and the fully connected layers in the policy network $\pi$ can be effectively pre-trained by learning to solve a proxy task that does not require knowledge about the particular layout or structure of a map or the probabilistic model governing the object placement. For this proxy task we randomly generate maps and place objects from $\cC^{\text{map}}$ and $\cC^{\text{targets}}$ in it. It is important to understand that the object placement is \emph{not} controlled by the same probabilistic model that is used to construct the graph maps for the actual navigation tasks. We then train the network to solve a classification task: given a class representation $\vy_{\text{target}}$, which pose nodes are connected to an instance of this class? We randomly sample a different target class for every minibatch and use a binary cross entropy loss function. Pre-training on this proxy task initialises the weights in all network layers to values that are useful to interpret the semantic word vector representations. We demonstrate in Section~\ref{sec:speedup} that this speeds up the learning process for the actual navigation task we are interested in.

\section{Evaluation}
We evaluate all 200 trained agents on 1000 episodes each: we evaluate on 100 randomly generated maps per agent, using 20 different probabilistic models $\cP$ controlling the target object locations, and 10 randomly chosen target objects per map. 

For each episode, a random target object class is chosen, such that at least one instance of such an object exists in the map. The agent is given a time budget of 10 time steps, i.e. the agent can sample a maximum of 10 goal locations to navigate to. As during training, we assume the agent can always reach its selected goal location, and do not take geodesic distance into account. If an object of the target class is connected to the selected goal location, the episode ends successfully. Otherwise, the agent samples the next goal from the policy until the 10 time steps are used up, in which case the episode ends unsuccessfully.

\subsection{Baselines} 
We compare the learned policies against two baselines:
\paragraph{Random Policy} The random policy baseline chooses a random pose node in the graph as the navigation goal. It never chooses the same node twice. 

\paragraph{Oracle Policy} The oracle policy has full access to the probabilistic model that controls where target objects appear in the environment. It can calculate the probability of finding the target object at any pose node, and chooses the node with the maximum probability as the navigation target. If the target object cannot be found there, it chooses the next likely goal and so on. It never chooses the same goal twice.

\subsection{Performance Metrics}
We use the following metrics to characterise performance:

\paragraph{Success Rate} How often can the agent find the target object within the available budget of 10 time steps? The agent can navigate to 10 goal locations before the episode ends unsuccessfully. 
\paragraph{Steps to Target} To how many goals does the agent navigate to on average before it eventually finds the target object? Notice that this measure only incorporates \emph{successful} episodes.


\begin{figure}[t]
    \centering
    \includegraphics[width=0.48\linewidth]{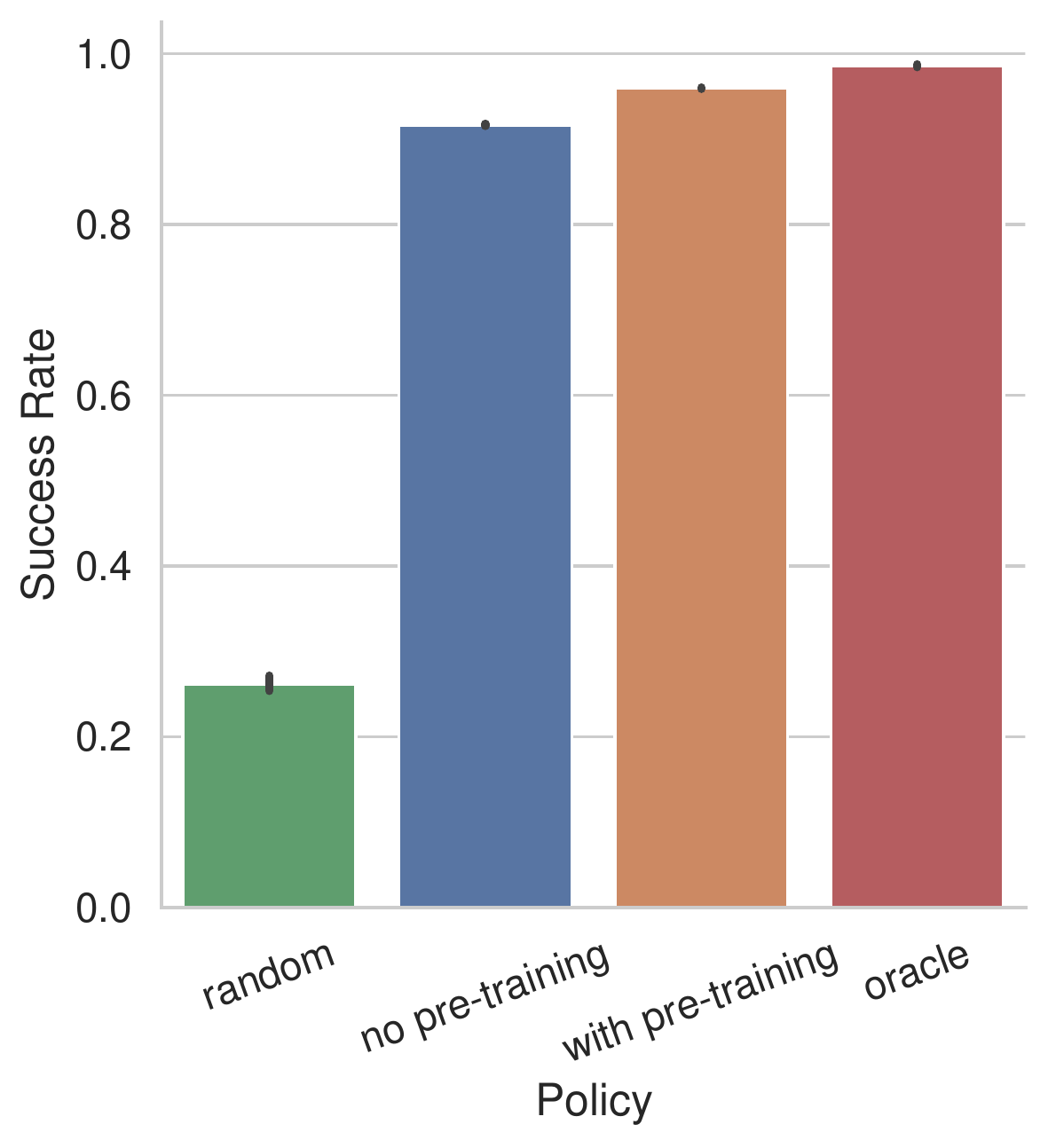}
    \includegraphics[width=0.48\linewidth]{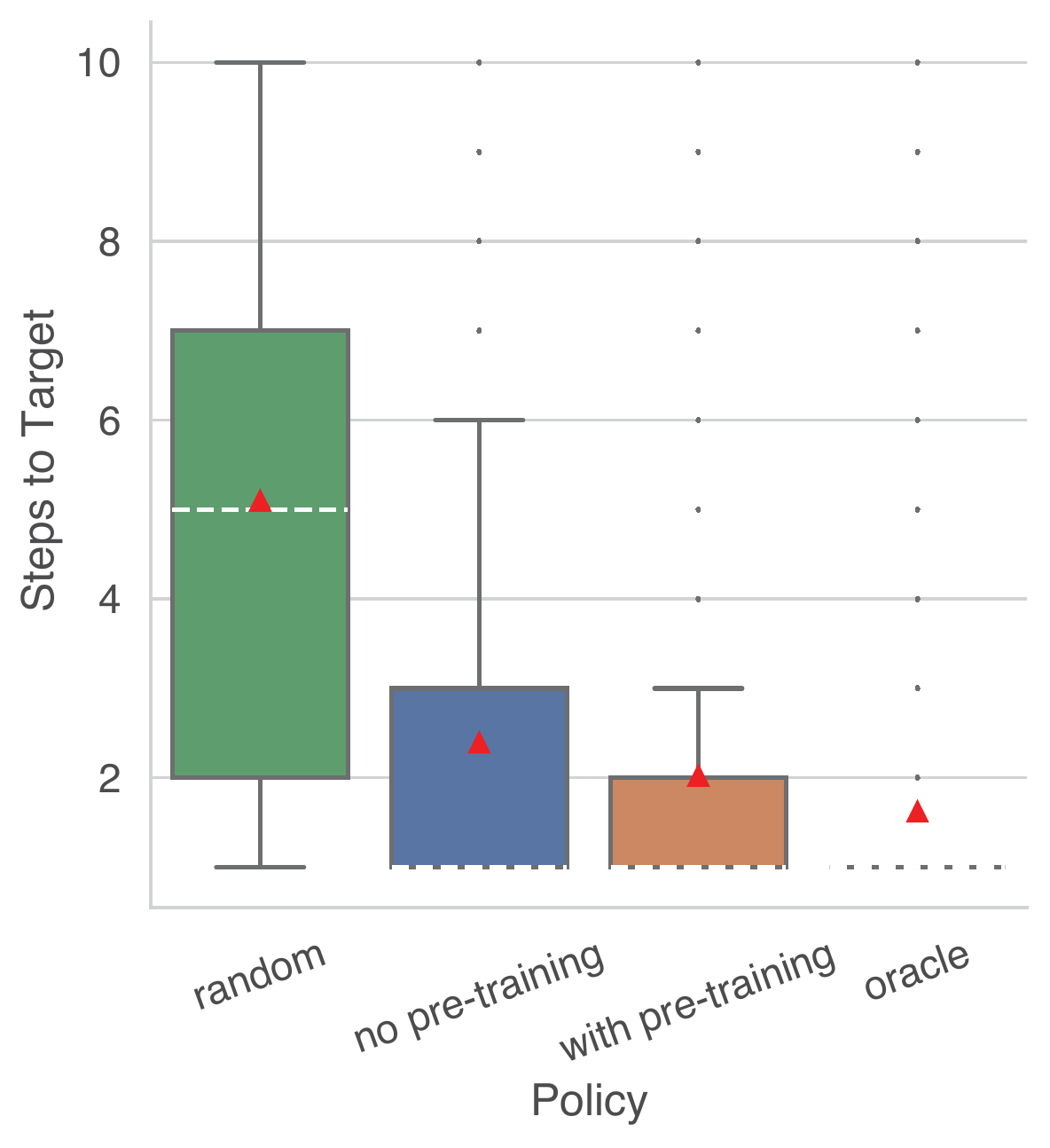}
    \caption{Success rate (left) and distribution of steps to target (right) combined over all target classes for different policies. Right plot: the median is represented by a white dashed line, the mean by a red triangle.  Results aggregated over 200,000 evaluation episodes: 20 probabilistic models $\times$ 10 agents $\times$ 100 map layouts $\times$ 10 randomly chosen target objects. Neither the environments nor the probabilistic models controlling the object placement have been used during training.
    }
    \label{fig:eval_success}
\end{figure}

\begin{table*}[bt]
\caption{Performance of different policies on the training environments and novel environments not encountered in training. We measure performance using success rate (first row) and required steps to target (second row).}
\label{tab:results}
\vskip 0.15in
\begin{small}
    \centering
    \begin{tabular}{@{}lcccc|cccc@{}}
        \toprule
         & \multicolumn{4}{c}{Evaluate on Training Environment}  & \multicolumn{4}{c}{Evaluate on Unseen Environments} \\ \midrule
         & random & no & with & & random & no & with & \\
         & policy & pre-training & pre-training & oracle  & policy & pre-training & pre-training & oracle \\ \midrule
         success r. & $ 0.33 \pm 0.47 $ & $ 0.98 \pm 0.13 $ & $ 0.99 \pm 0.09 $ & $ 0.99 \pm 0.09 $ & $ 0.26 \pm 0.44 $ & $ 0.92 \pm 0.28 $ & $ 0.96 \pm 0.20 $ & $ 0.99 \pm 0.12 $ \\
         steps to t.& $ 5.00 \pm 2.70 $ & $ 1.41 \pm 1.03 $ & $ 1.45 \pm 1.09 $ & $ 1.66 \pm 1.51 $ & $ 5.10 \pm 2.89 $ & $ 2.39 \pm 2.07 $ & $ 2.02 \pm 1.78 $ & $ 1.62 \pm 1.49 $ \\
         \bottomrule
    \end{tabular}
\end{small}
\vskip -0.1in
\end{table*}

\begin{figure*}[t]
    \centering
    \includegraphics[width=0.32\linewidth]{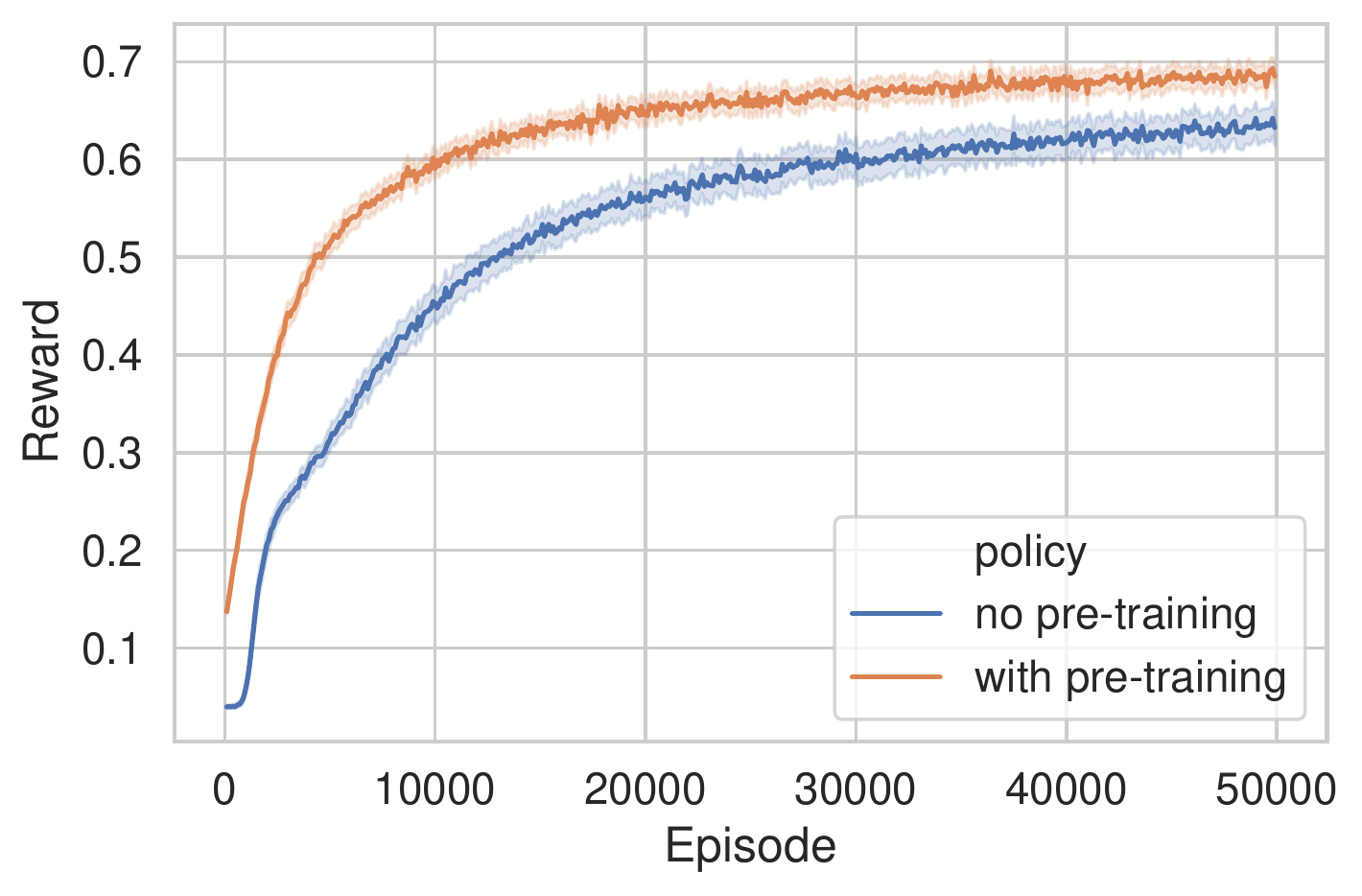}
    \includegraphics[width=0.32\linewidth]{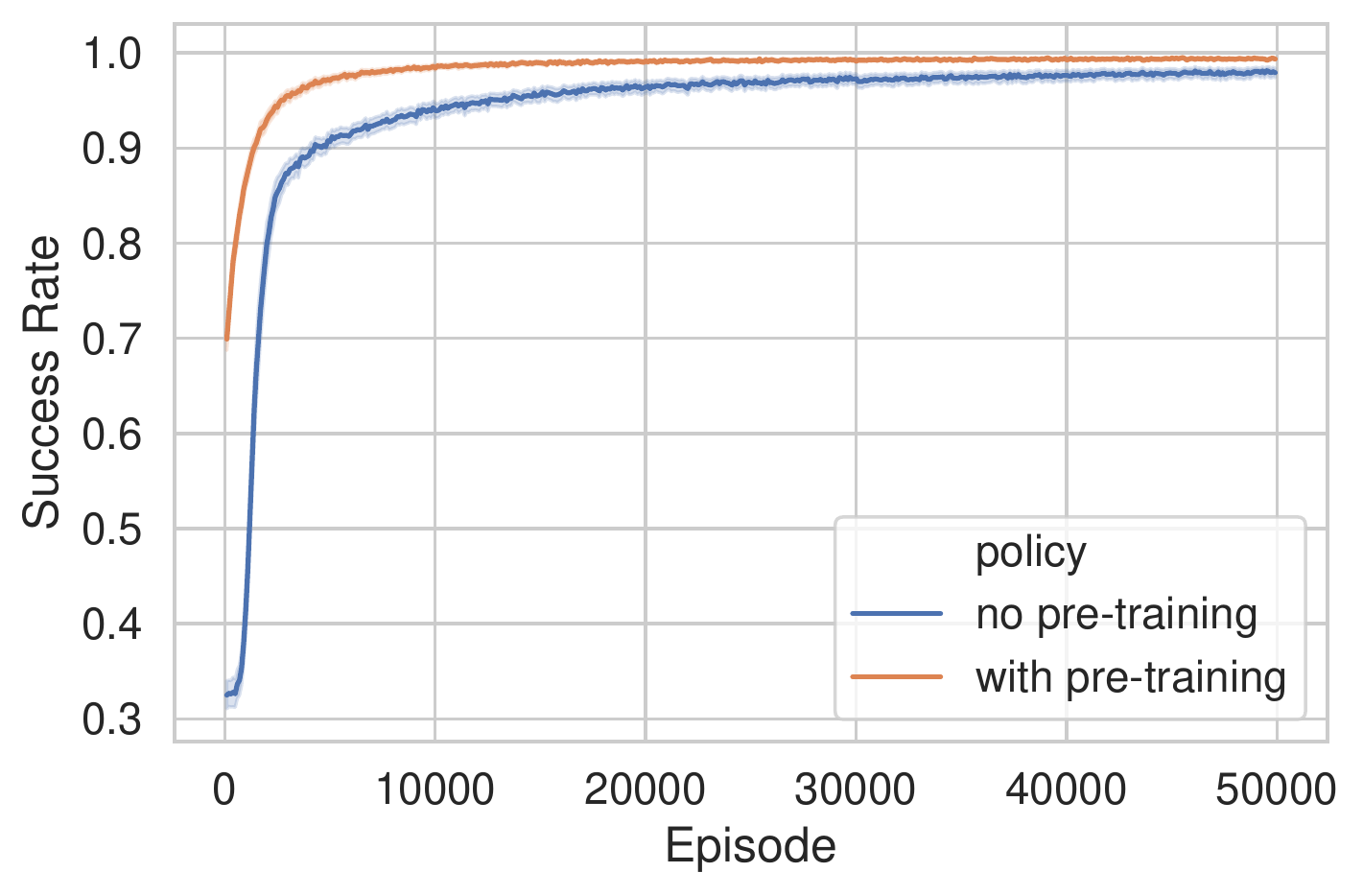}
    \includegraphics[width=0.32\linewidth]{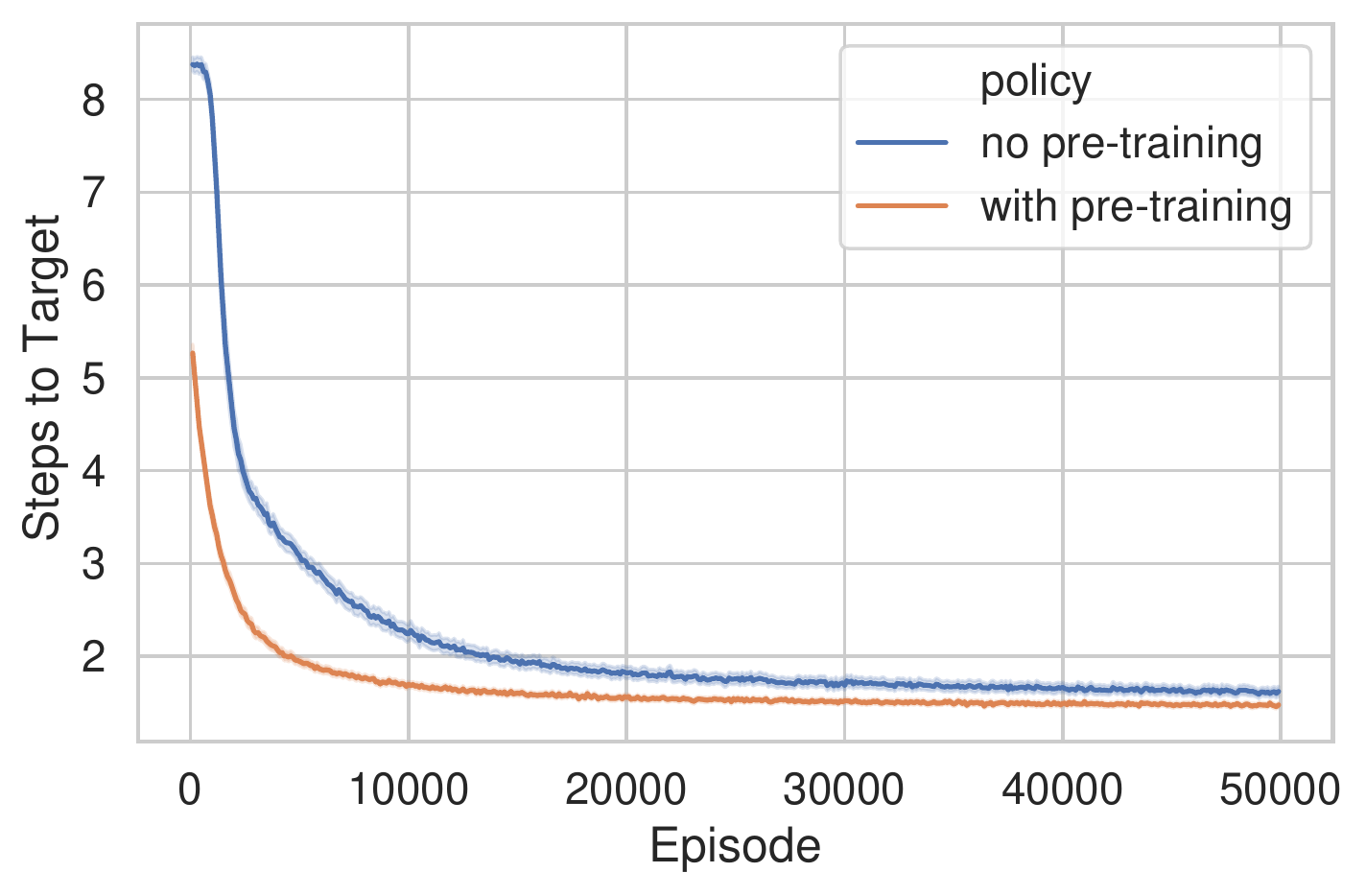}
    \caption{Reward (left), success rate (centre), and steps to target (right) averaged over 200 training runs (10 randomly initialised networks $\times$ 20 environments with different probabilistic model). The shaded region around the line corresponds to the 90th percentile. When the policy network is initialised by the proxy task pre-training (explained in Section~\ref{sec:pretrain}), it learns significantly faster, reaching the same level of performance after a fraction of the training episodes.}
    \label{fig:training}
\end{figure*}

\begin{figure}[t]
    \centering
    \includegraphics[width=\linewidth]{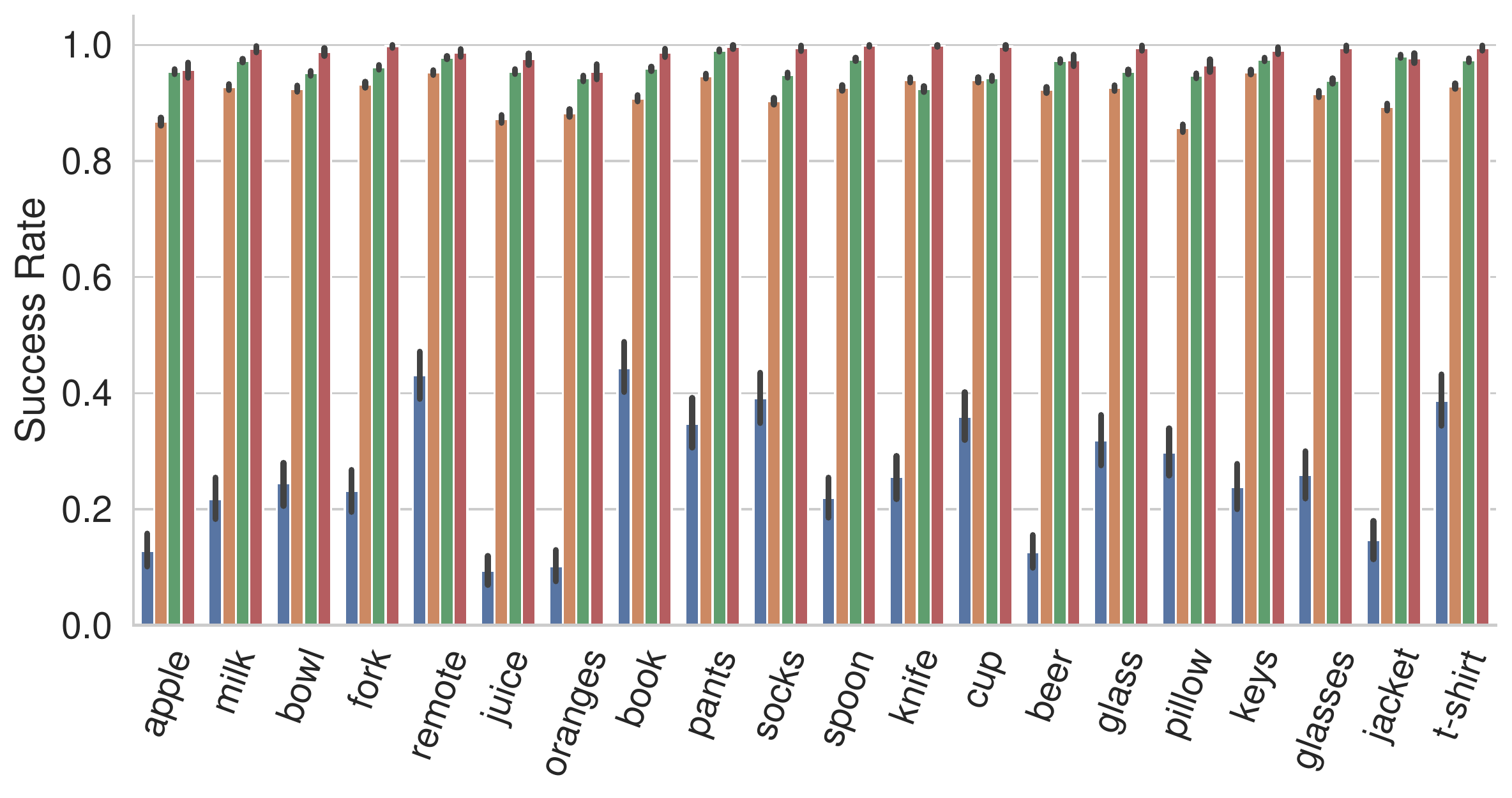}
    \includegraphics[width=\linewidth]{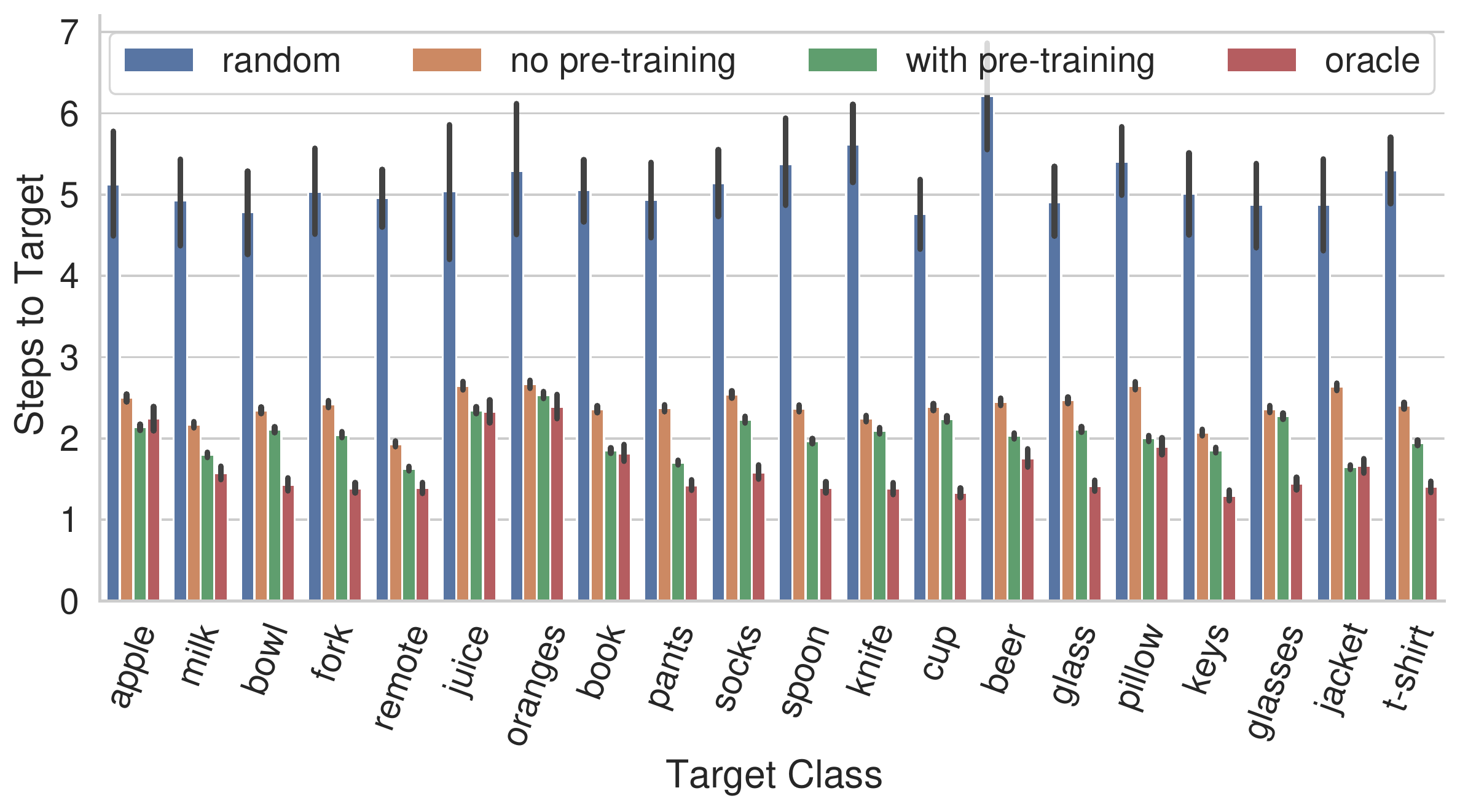}
    \caption{Class-wise success rates (top) and steps to target (bottom) for different policies. Results aggregated over 200,000 evaluation episodes. 
    }
    \label{fig:seen_classes}
\end{figure}

\begin{figure}[t]
    \centering
    \includegraphics[width=\linewidth]{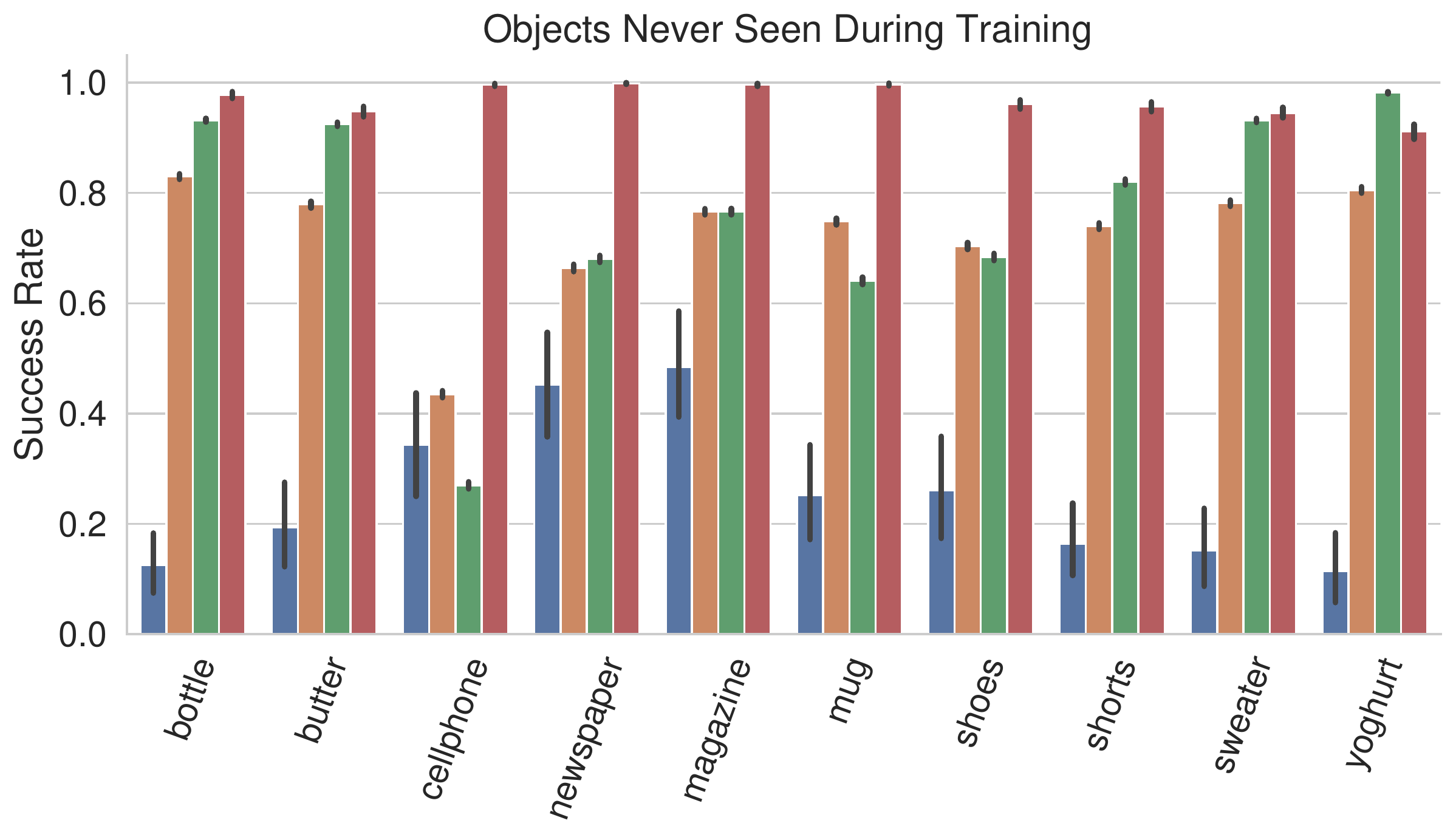}
    \includegraphics[width=\linewidth]{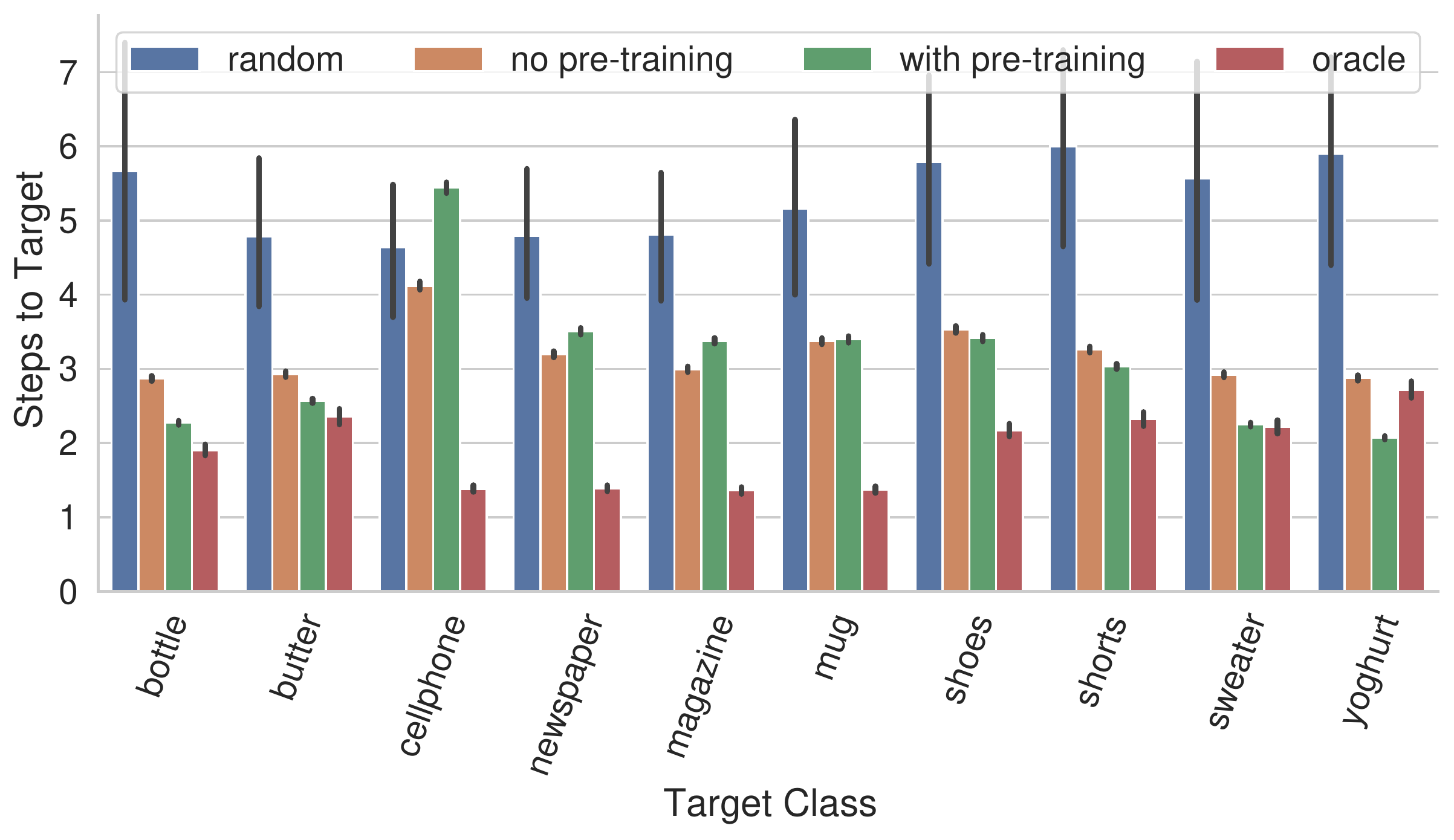}
    \caption{Class-wise success rates (top) and steps to target (bottom) for different policies and target objects that were never seen during training. The learned policies generalise to these unseen objects as long as they are semantically close and behave similarly as objects the policy was trained on. The \emph{cellphone} class is an exception here, since none of the original training classes is semantically close to it.}
    \label{fig:unseen_classes}
\end{figure}

\begin{table}[bt]
\caption{Results on unseen environments with \emph{unseen} target objects. Success rate (first row) and steps to target (second row).}
\label{tab:results_unseen}
\vskip 0.15in
\begin{small}
    \centering
    \begin{tabular}{@{}lcccc@{}}
        \toprule
         & random & no & with & \\
         & policy & pre-training & pre-training & oracle \\ \midrule
         succ. \hspace{-1 em} & $ 0.25 \pm 0.43 $ & $ 0.72 \pm 0.45 $ & $ 0.76 \pm 0.43 $ & $ 0.97 \pm 0.17 $ \\
         steps & $ 5.14 \pm 2.99 $ & $ 3.16 \pm 2.54 $ & $ 2.90 \pm 2.44 $ & $ 1.89 \pm 1.78 $ \\
         \bottomrule
    \end{tabular}
\end{small}
\end{table}

\section{Results}
This section explains the four key results and insights gained from the conducted experiments.

\subsection{Graph Convolutional Networks Can Successfully Learn Object-centric Navigation Policies on Semantic Maps}
The left side of Table~\ref{tab:results} compares the performance of two learned policies with the two baseline policies (random and oracle). As we can see, the policy networks successfully learned to find target objects in their training environments. The trained policies achieve almost perfect results, finding the target object after around $1.4$ steps on average and successfully finishing 99\% of the episodes in time (with pre-training, 98\% for the policy without pre-training). The random policy only finishes 33\% of all episodes successfully (i.e. it eventually finds the target object within 10 time steps) and takes 5 steps on average if successful.
The oracle policy that has full access to the probabilistic model that controls where objects tend to appear in the environment achieves 99\% success rate with 1.66 steps to target on average. 

\subsection{Pre-Training with a Proxy Task Speeds Up Learning}
\label{sec:speedup}
The proxy task procedure explained in Section~\ref{sec:pretrain} significantly speeds up the training process. When the policy network is initialised in this way, it reaches the same performance as the uninitialised network after a fraction of the training episodes. This can be clearly observed in the plots of Fig.~\ref{fig:training} that illustrate how reward, success rate, and steps to target evolve during training. The results in these plots are averages (with the 90th percentile represented by the semi-transparent area) over the behaviour of 200 training runs (20 environments, training 10 agents each).

\subsection{The Learned Policy Generalises to Unseen Environments}
\label{sec:generalise_unseen_environment}
Table~\ref{tab:results} compares the average performance of different policies when evaluated in their training environment and when transferred into a novel, unseen environment with different characteristics. 

When transferred into unseen environments with new randomised probabilistic characteristics, the performance drops only slightly from 99\% to 96\% success rate and from 1.4 to 2.0 steps to target on average (from 98\% to 92\%, and from 1.4 to 2.39 steps without pre-training). Thus we can conclude that the learned policy (especially when initialised with proxy task pre-training), generalises well to unseen environments where the placement of potential target objects is controlled by a different hidden probabilistic model. Notice that although the probabilities for spawning target objects at certain map objects will generally be different between training and the testing environment (i.e. $\cP^{\text{train}}(c_{\text{target}} | c_{\text{map}}) \neq \cP^{\text{test}}(c_{\text{target}} | c_{\text{map}})$), the possible pairings of which target objects appear near which map objects, remains unchanged. That is, as listed in footnote~\ref{note:spawn}, a \emph{book} might appear next to either \emph{bedside}, \emph{shelf}, or \emph{sofa}, but not next to other map objects such as \emph{kitchen-table}. The policy network is able to learn these underlying characteristics that are constant\footnote{An interesting experiment for future work is to investigate to what extend the policy network can learn to adopt to situations where these characteristics are variable between environments.} between different environments and thus generalise to different environments with a different map layout (i.e. graph connections, room layout etc) and different $\cP$.

Fig.~\ref{fig:eval_success} illustrates the different success rates and the distribution of the required steps to target for the different policies in the unseen environments. As can be seen in the boxplot of Fig.~\ref{fig:eval_success} (right), while the median of both learned policies and the oracle policy is 1, the overall distribution differs substantially. The policy network that is initialised with pre-training has a smaller spread than the policy that is randomly initialised before training. The policies initialised with the pre-trained network also require less steps to target on average (red triangles).
Fig.~\ref{fig:seen_classes} illustrates the performance for all target classes individually. 
For comparison, the oracle consistently finds the target after 1.6 steps on average, with a success rate of 99\%, while the random policy takes an average of 5 steps and only concludes around 30\% of episodes successfully.

\subsection{The Learned Policy Generalises to Unseen Classes}
\label{sec:generalise_unseen_objects}
How well can the learned policies generalise to target classes that were never encountered during training (or pre-training)? To answer this question, we removed all known target objects and replaced them with unknown objects that are in most cases semantically similar to one of the known classes, and tend to appear in similar places.

As illustrated in Fig.~\ref{fig:unseen_classes}, the policies generalise well. This performance is based on the expressive power of the FastText word vectors that capture semantic similarity. With the exception of the \emph{cellphone} class that did not have a semantically similar class in the training dataset, the performance of both learned policies is much better than random, and close to the oracle policy. As before, agents using a policy network that was initialised with proxy task pre-training tend to perform better. Table~\ref{tab:results_unseen} summarises the results.



\section{Conclusions}
The recently emerging class of object-based semantic SLAM systems provide us with very compact, yet rich representations of the environment of mobile robots. Especially applications such as domestic service robotics and elderly care robotics can benefit from the graph-based maps that contain rich semantic information. However, more research into how the maps generated by this new class of SLAM systems can be used most beneficially is needed. 

We contributed to this exciting new direction of research and demonstrated that a graph convolutional network is able to learn a navigation policy on such a graph. We have shown that word vectors are useful representations for landmark classes in this context, allowing the navigation policy to generalise to semantically similar, but previously unseen object categories.
In future work we hope to evaluate this approach online on a robot, in concert with an object-based semantic SLAM sytem such as QuadricSLAM~\cite{nicholson2018quadric}, Fusion++~\cite{McCormac2018} or others~\cite{hosseinzadeh2019real}. We have not yet studied the influence of various hyperparameters, the network architecture, or the word vector representation on the performance of the presented approach. 

\bibliography{bibfile.bib}
\bibliographystyle{icml2020}

\end{document}